\documentclass{article}
\usepackage{ijcai26}

\usepackage{times}
\usepackage{soul}
\usepackage{url}
\usepackage[hidelinks]{hyperref}
\usepackage[utf8]{inputenc}
\usepackage[small]{caption}
\usepackage{graphicx}
\usepackage{amsmath}
\usepackage{amsthm}
\usepackage{booktabs}
\usepackage{algorithm}
\usepackage{algorithmic}
\usepackage[switch]{lineno}
\usepackage{multirow}
\usepackage{amsmath, amssymb}
\usepackage{bm}
\usepackage[T1]{fontenc}
\usepackage[utf8]{inputenc}
\usepackage{lmodern}

\title{Logi-PAR: Logic-Infused Patient Activity Recognition via Differentiable Rule}

\author{
    Muhammad Zarar, MingZheng Zhang, Xiaowang Zhang, Zhiyong Feng, Sofonias Yitagesu, Kawsar Farooq
    \affiliations
    School of Computer Software, Tianjin University, China
    \emails
}

\begin{document}

\maketitle

\begin{abstract}
Patient Activity Recognition (PAR) in clinical settings uses activity data to improve safety and quality of care. Although significant progress has been made, current models mainly identify \textit{which} activity is occurring. They often spatially compose sub-sparse visual cues using global and local attention mechanisms, yet only learn logically implicit patterns due to their neural-pipeline. Advancing clinical safety requires methods that can infer \textbf{why} a set of visual cues implies a risk, and how these can be compositionally reasoned through explicit logic beyond mere classification. To address this, we proposed \textbf{Logi-PAR}, the \textit{first Logic-Infused Patient Activity Recognition Framework} that integrates contextual fact fusion as a multi-view primitive extractor and injects neural-guided differentiable rules. Our method automatically learns rules from visual cues, optimizing them end-to-end while enabling the implicit emergence patterns to be explicitly labeled during training. To the best of our knowledge, Logi-PAR is the first framework to recognize patient activity by applying learnable logic rules to symbolic mappings. It produces auditable \textit{why} explanations as rule traces and supports counterfactual interventions (e.g., risk would decrease by 65\% if assistance were present). Extensive evaluation on clinical benchmarks (VAST and OmniFall) demonstrates state-of-the-art performance, significantly outperforming Vision-Language Models and transformer baselines. The code is avliable via\url{https://github.com/zararkhan985/Logi-PAR.git}.
\end{abstract}

\section{Introduction}
Patient Activity Recognition (PAR) from clinical imagery has emerged as a cornerstone of modern hospital safety, increasingly deployed to provide early risk for critical events such as unsafe bed exits, falls, and mobility deterioration. However, unlike standard action classification tasks (e.g., labelling "sitting" vs. "walking"), clinical PAR faces unique challenges that make it fundamentally more difficult. The decisive cues for patient activity risk assessment are often sparse, rare, and deeply relational, such as the precise state of a bed rail, the proximity of a patient's pelvis to the bed edge, or subtle contact with a caregiver. These fine-grained signals may occupy only a fraction of the image, appear in only one camera view, or be occluded or obscured by poor lighting. Consequently, reliable patient safety monitoring requires more than merely predicting which activity is occurring; it requires a system capable of inferring "why" a specific configuration of these subtle cues constitutes a high-risk state, providing actionable information and explanations essential for decision-making.

\begin{figure}[t] 
    \centering
    \includegraphics[width=\linewidth]{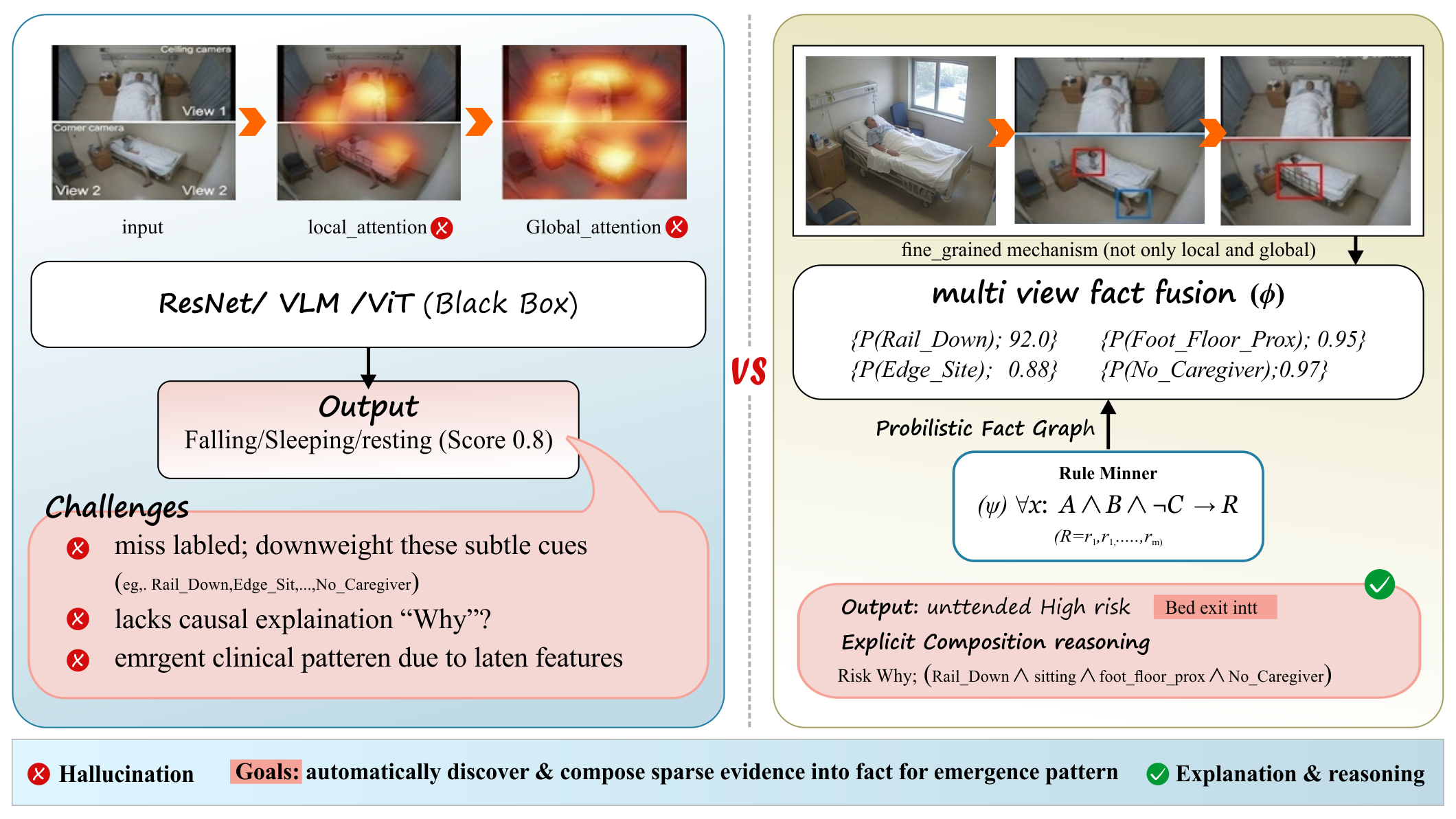}
    \caption{\textbf{Overcoming Black-Box Correlations to Logical Inference in Clinical Risk Detection}. Standard attention mechanisms frequently miss decisive but sparse risk cues, mistaking high-risk exits for resting states due to background bias. Logi-PAR overcomes this by enforcing a hierarchical inference structure: it first extracts reliable \textit{Atomic Facts} via fine-grained attention and then applies learnable logic rules. This explicitly models the causal structure of an \textit{Unattended Bed Exit}, ensuring correct classification and generating human-verifiable explanations.}
    \label{fig:motivation}
\end{figure}

Despite substantial progress in state-of-the-art (SOTA) models, including large Vision-Language Models (VLMs) like GPT-4V and MedViT, which optimize for human activity in PAR, average classification performance remains low across global scene statistics. In clinical settings, decisive risk cues are often sparse, fine-grained, and relational, occupying small image regions (e.g., bed-rail position, foot-floor proximity, or hand-support contact) rather than a single dominant visual pattern. The existing SOTA attention mechanisms often downweight these subtle cues in favour of dominant background features, leading to failures in ambiguous or rare compositional cases. Figure \ref{fig:motivation} highlights an example of a benchmark where Figure. \ref{fig:motivation} (a) shows existing SOTA models rely on global attention heatmaps that often miss decisive but sparse visual cues—such as a lowered bed rail or specific foot positioning leading to misclassifications like labelling a high-risk bed-exit attempt as safe "sleeping." This challenge is becoming critical for the PAR setting, where relational cues are subtle, scattered, and partially missing. They need to introduce a rule-based paradigm that bridges this gap by shifting from holistic pattern matching to compositional reasoning, treating distinct visual evidence (eg; "Rail Down," "No Caregiver") as atomic fact pieces to be discovered and assembled. This goal shifts to explicit compositional reasoning, such as “What" is happening, "why" it's happening, and reasoning about what new patterns exist.

It is known that the logical rule can be used to extract sparse cue evidence by inferring a compositional logic mechanism to enhance the patient's activity-understanding by filling in missing atomic facts and rare cues. Figure \ref{fig:motivation} (b) demonstrates the desired outcome of inferring the missing atomic fact from pixel-precise correlation by a differentiable logic rule. The framework explicitly detects these fine-grained evidences and logically combines them (e.g., Rail Down AND Foot-Floor Proximity AND No Caregiver) to correctly identify and explain complex, emergent PAR clinical states, such as "Unattended High-Risk Bed Exit Intent." Thereby providing the transparent, causal explanation towards "why" reasoning that is currently missing from black-box approaches. These limitations reflect fundamental gaps in current SOTA PAR approaches, including ClinActNet~\cite{majid2025multimodal}, VLM-AR~\cite {abid2025vision}, and NSKI~\cite {magnini2025neuro}. In more detail, ClinActNet is the first multimodal LLM method that integrates different input modalities to improve the recognition of complex activities under clinical variability.
In contrast, VLMs-AR enhance HAR to a health-monitoring state, highlighting the flexibility of generative VLMs, and propose a caption-style dataset and evaluation methodology to account for the non-deterministic nature of VLM outputs. Subsequently, NSKI focuses on chronic disease diagnosis, combining symbolic reasoning with learned components to improve interpretability and robustness. Meanwhile, it primarily operates on structured clinical variables rather than extracting compositional evidence directly from images. Although these SOTA baselines have shown promising performance for activity recognition in patient monitoring, they suffer from implicit logical constraints that hinder transparent reasoning and fail to capture fact evidence, and lack a mechanism to convert atomic actions into emergent patterns for PAR. The goal is not only to achieve black-box classification of data that were never explicitly labelled as single classes, but also to answer not only what is happening but also why it constitutes a clinical state, through explicit logic rules.

We propose \textbf{Logi-PAR}, a logic-infused PAR framework that explicitly represents clinical evidence as probabilistic atomic facts and learns a compact set of differentiable logic rules. Given synchronized multi-view images, Logi-PAR first predicts calibrated predicate probabilities from the backbone logits (e.g., \textsc{RailDown}, \textsc{EdgeSit}, \textsc{SupportContact}, \textsc{CaregiverNear}) and fuses them into an uncertainty-aware \emph{atomic-fact graph}. A neural-guided rule learner then induces soft logical rules (including negation) that compose these facts into activity/state decisions.
This yields: menaced to unseen patient activity cues-combinations (compositional generalization) beyond labelled categories, risk states as reusable facts, rather than memorizing entire patterns. Secondly, risk states are compositions of reusable fact actionable  "why" explanations that can be validated as rule chains and (iii) tested via counterfactual predicate interventions for false alarm rate.

Our key contributions are:
\begin{itemize}
    \item We proposed the first logic inference framework, \textbf{Logi-PAR}, that jointly learns rule structures and visual fact grounding in an end-to-end, differentiable pipeline. 
    \item We propose a neural-guided compositional rule learner that automatically discovers soft logic rules from atomic facts, with confidence estimates for the factual spatial ambiguity in clinical settings. 
    \item To introduce multi-view fact fusion, transforming visual cues into a probabilistic scene activity graph of atomic predicates rather than entangled global features.
    \item To evaluate the SOTA benchmark that explicitly tests generating "why" a specific configuration of these subtle cues constitutes a high-risk state that provides an actionable explanation for PAR.
\end{itemize}

\section{Related Work}
\subsection{Patient Activity Recognition(PAR)} 
Computer vision for continuous patient monitoring has evolved from simple motion detection to scalable safety systems capable of operating in complex hospital environments. Recent advancements focus on real-time video analysis pipelines and the release of large-scale, annotated datasets that capture the "long tail" of patient behaviors in realistic care settings \cite{yuan2024self} \cite{lookdeep2025monitoring}. For instance, the CASTLE benchmark introduces multi-perspective ego-centric and exo-centric views to better capture fine-grained interactions \cite{rossetto2025castle}, whereas other efforts target specific clinical sub-problems, such as in-bed segmentation and pose estimation under occlusion \cite{mennella2025advancing}. Fall detection remains a central challenge, with recent benchmarks such as SPLAT-Flow \cite{go2025splatflow} and multi-view protocols \cite{wang2024toward} \cite{zhou2024survey} aiming to assess generalization from staged simulations to the unstructured reality of hospital rooms.

Despite these strides, purely data-driven PAR methods remain brittle in the presence of severe domain shifts, lighting variations, and occlusions inherent to clinical deployment. Recent studies in unsupervised video domain adaptation, such as TranSVAE \cite{wei2023uvda_disentanglement} and UNITE  \cite{reddy2024unsupervised}, highlight that statistical alignment alone often fails to preserve semantic consistency across diverse hospital layouts. This limitation motivates Logi-PAR, which moves beyond pixel-level statistical features to leverage symbolic predicates. There is grounding recognition in logical definitions (e.g., spatial relationships that define a "fall") rather than texture or background cues. Logi-PAR aims to achieve robustness to visual-domain shifts in the patient activity state.

\subsection{Vision Language Models for PAR} 
The integration of Vision-Language Models (VLMs) has rapidly expanded the scope of patient monitoring, enabling systems to process multimodal inputs and generate descriptive activity summaries. Highly capable omni-models and medical-specific adaptations, such as Med-Gemini and recent variant methods \cite{li2025visionunite}, \cite{jiang2025omniv} \cite{MedLogic-AQA}, and \cite{nath2025vila} , have demonstrated impressive performance on standard visual question answering (VQA) tasks. These models promise a holistic understanding of patient state by synthesizing visual data with clinical context. However, their deployment is severely hampered by the "black box" nature of their reasoning and a propensity for hallucination—generating plausible but factually incorrect descriptions of patient actions or clinical patterns.

Similarly, various STOA evaluation benchmarks, such as VIDHALLUC \cite{li2025vidhalluc} and MedHallu \cite{pandit2025medhallu}, show that hallucinations in Large Multimodal Models (LMMs) persist across temporal reasoning, multi-object tracking, and free-form generation. These failures are critical in clinical settings where "explanations" must be verifiable and safety-critical \cite{zhang2025EMRRG} \cite{majid2025harmonizing}. Consequently, Logi-PAR rejects the reliance on unconstrained, probabilistic language generation. Instead, it targets \emph{structured} explanations via explicit predicates and rule chains, ensuring that every system output is grounded in verifiable visual evidence and subject to auditable failure analysis.

\subsection{Rule Learning for PAR} 
To combine perceptual learning with robust reasoning, differentiable neuro-symbolic methods have gained renewed attention. Recent approaches such as Neural Rule Learner (NeurRL) \cite{gao2026differentiable} and NeSyCoCo \cite{kamali2025nesycoco} have made significant progress in learning logic programs directly from raw data and in enhancing compositional generalization for PAR. Parallel advancements in differentiable Inductive Logic Programming (ILP) \cite{shindo2024learning} and logic-guided verification frameworks, such as NeuS-QA \cite{shah2025neusqa}, demonstrate that embedding temporal logic constraints into neural networks can enforce consistency and improve long-horizon reasoning \cite{fitas2025neuro} \cite{xiong2024tilp}.

However, these methods often struggle to learn clinically meaningful compositional rules from the noisy, probabilistic atomic facts extracted from multi-view clinical imagery. Furthermore, standard evaluation protocols rarely test for counterfactual robustness or performance on unseen combinations of known primitives—a frequent occurrence in complex patient trajectories \cite{Zhao_2025_CVPR}. Despite this, Logi-PAR addresses it by introducing a differentiable rule-induction framework designed explicitly for probabilistic clinical fact graphs. It validates these rules under rigorous compositional and counterfactual protocols, aligning the evaluation with the intended "why" reasoning behavior required for trusted clinical decision support.

\section{Problem Formulation}
Let $\mathcal{I} = \{I_1, I_2, \dots, I_V\}$ represent the input space, a set of synchronized clinical images captured from $V$ distinct camera views at a given timestamp, and the output space. Our goal is to infer a clinical risk state $y \in \mathcal{Y}$ (eg, Unattended Bed Exit Intent ) while simultaneously generating an explicit explanation $\mathcal{E}_{e}$ and enabling counterfactual reasoning $\mathcal{E}_{cf}$.

To bridge the gap between pixel-level inputs and high-level risk semantics, we define the following symbolic components:

\textbf{Atomic Facts (Predicates):} We define a vocabulary of semantic primitives $\mathcal{P}=\{p_1, p_2,\dots, p_n\}$, where each $p_i$ represents a specific, clinically meaningful atomic observation (e.g., \texttt{rail\_down}, \texttt{edge\_sitting}, \texttt{caregiver\_present}). Unlike end-to-end PAR models that entangle such cues in latent representations, rather than explicitly as a probabilistic interface between perception and reasoning.

\textbf{Fact Confidence \& Attribution:} In a given input $\mathcal{I}$, the system estimates a confidence score $c_i \in [0,1]$ for each atomic fact $p_i$, alongside a view attribution vector $\mathbf{v}_i \in \{0,1\}^V$ which identifies the subset of views providing evidence for that fact. This view attribution is essential for multi-view clinical imaging, where critical evidence (such as rail state, foot placement, caregiver contact) may be visible in only one view or partially occluded in others.

\textbf{Compositional Rules:} We define a set of logical rules
$\mathcal{R} = \{r_1, r_2, \dots, r_M\}$, each rule $r_j$ functions as a compositional operator that aggregates a specific configuration of atomic facts to imply a clinical state (or intermediate state). Critically, rules may include negation to represent clinically meaningful “absence” conditions (e.g., risk is high only if assistance is absent).

\textbf{Factorize Mapping} The framework learns a composite mapping function that factorizes perception and reasoning:
\begin{equation}
    \mathcal{I} \xrightarrow{\phi} \mathcal{G}(\mathcal{P}, \mathbf{c}, \mathbf{v})
    \xrightarrow{\psi} \mathcal{R} \to (y, \mathcal{E}_{e}, \mathcal{E}_{cf})
\end{equation}

where $\phi$ is the perception module that maps raw images to a probabilistic fact graph $\mathcal{G}$, and $\psi$ is the differentiable reasoning module that applies rules $\mathcal{R}$ to infer the final state $y$ and generate causal traces.

\section{LogiPAR}
\subsection{Overview}
As illustrated in Figure \ref{eq:framework_overview}, we propose \textbf{Logi-PAR}, a logic-infused framework designed to learn a composite mapping $\mathcal{F}: \mathcal{I} \rightarrow (y, \mathcal{E}_{e}, \mathcal{E}_{cf})$. Unlike black-box networks that directly map images to labels, we formally factorize this mapping into two coupled differentiable stages: 

\noindent\textbf{Multi-View Primitive Factorization (Perception $\phi$):} Converts clinical images into a set of patient-state ``atomic facts'' with calibrated uncertainty.

\noindent\textbf{Neural-Guided Differentiable Logic (Reasoning $\psi$):} Learns a compact set of compositional logic rules (including negation) to map atomic facts to decisions and output explicit rule traces.

This decomposition is motivated by the observation that clinically significant states rarely manifest as a single visual pattern but rather emerge from the composition of multiple subtle cues. This architecture allows the model to explicitly model the causal path from pixels to facts, and finally to decisions. The end-to-end forward propagation is formally defined as:
\begin{equation}
\label{eq:framework_overview}
    \hat{y}, \mathcal{E}_{e}, \mathcal{E}_{cf} = \psi \circ \phi(\mathcal{I}) \implies \mathcal{I} \xrightarrow{\phi} \mathcal{G} \xrightarrow{\psi} (y, \mathcal{E}_{e}, \mathcal{E}_{cf})
\end{equation}

\noindent where the intermediate representation $\mathcal{G}$ is the generated \textbf{probabilistic fact graph}, comprising the atomic fact confidence vector $\mathbf{c} \in [0,1]^N$ and the multi-view attribution matrix $\mathbf{v} \in [0,1]^{N \times V}$. Based on this formulation, Logi-PAR comprises three principal modules: (1) Multi-View Fact Fusion implementing $\phi$, (2) Neural-Guided Differentiable Rules implementing $\psi$, and (3) Causal explanatory output $\xi$.

\begin{figure*}[ht] 
    \centering
    \includegraphics[width=\linewidth]{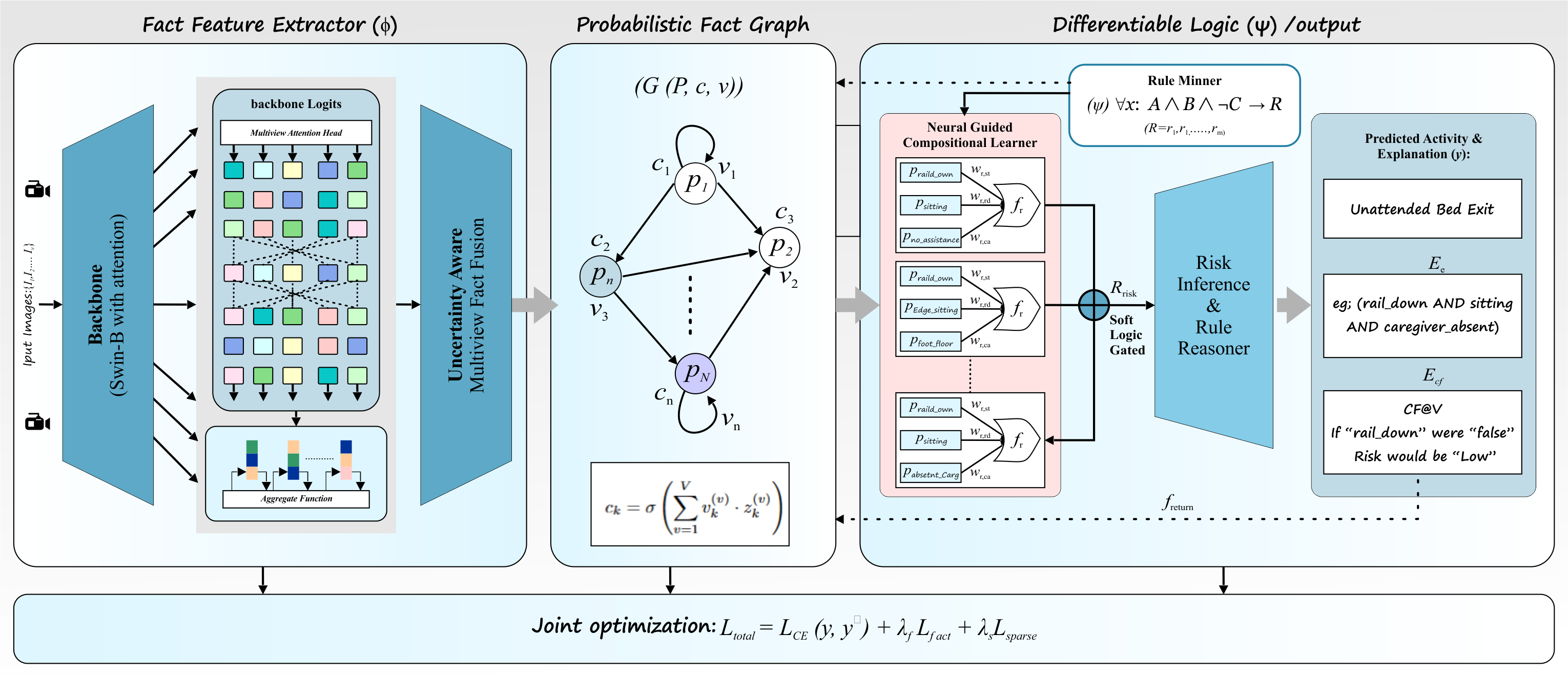}
    \caption{Overview of the proposed \textbf{Logi-PAR} framework. The framework processes multi-view images through a shared perception backbone to extract a Probabilistic Fact Graph. These facts are then fed into the Differentiable Causal-Logic Layer, where the Gated Soft-Logic Composition (Eq. 2) dynamically assembles atomic facts into complex clinical risk states, enabling both accurate classification and explanations.}
    \label{fig:framework}
\end{figure*}

\subsection{Multi-View Fact Fusion Perception}
The perception module $\phi$ is designed to extract robust atomic facts in the presence of dynamic occlusions and view inconsistencies. Unlike traditional feature level fusion, we propose an \textit{Uncertainty-Aware Logit Fusion} mechanism.

\noindent\textbf{View-Specific Prediction and Reliability.}
For input images $\{I^{(v)}\}_{v=1}^V$ from $V$ views, a shared backbone $f_\theta$ extracts feature maps $F^{(v)}$. Subsequently, parallel prediction heads output both the raw logit $z_k^{(v)}$ for the $k$-th atomic fact and a scalar reliability score $\rho_k^{(v)}$, representing the visibility of fact $k$ in view $v$:
\begin{equation}
\label{eq:view_pred}
z_k^{(v)} = h_k^{pred}(f_\theta(I^{(v)})), \quad \rho_k^{(v)} = h_k^{rel}(f_\theta(I^{(v)}))
\end{equation}

\noindent\textbf{View Attribution Weights.}
To handle clinical occlusions, we compute normalized view attribution weights $v_k^{(v)}$ based on the reliability scores. These weights not only facilitate weighted fusion but also serve as an auditable trace of evidence source:
\begin{equation}
\label{eq:view_attn}
v_k^{(v)} = \frac{\exp(\rho_k^{(v)})}{\sum_{u=1}^V \exp(\rho_k^{(u)}) + \epsilon}, \quad \text{s.t.} \sum_{v=1}^V v_k^{(v)} = 1
\end{equation}

\noindent\textbf{Uncertainty-Aware Fusion.}
The final confidence $c_k$ for each atomic fact is obtained via weighted logit pooling. This strategy ensures that high-reliability views dominate the decision while preserving uncertainty gradients for end-to-end training:
\begin{equation}
\label{eq:fusion}
c_k = \sigma\left( \sum_{v=1}^V v_k^{(v)} \cdot z_k^{(v)} \right)
\end{equation}

This completes the construction of the probabilistic fact graph $\mathcal{G}$, providing a semantic foundation for logical reasoning.

\subsection{Neural-Guided Differentiable Logic}
The reasoning module $\psi$ operates on the fact vector $\mathbf{c}$. To automatically discover sparse clinical risk patterns, we design a fully parameterized \textit{Neural-Guided Rule Learner} that jointly optimizes rule structure and parameters. The detailed interpretation is seen in \textbf{Appendix A.}

\noindent\textbf{Differentiable Literal Selection.}
The fundamental unit of a rule is the literal. For the $j$-th literal slot in the $m$-th rule, we introduce a learnable selection matrix $\mathbf{\Gamma}\_{m,j}\in\mathbb{R}^N$. Utilizing the Gumbel-Softmax technique, we generate a discretised selection distribution $\gamma_{m,j}$ to "softly select" the most relevant primitive from the pool of $N$ facts:
\begin{equation}
\label{eq:selection}
\gamma_{m,j} = \text{Softmax}\left(\frac{\mathbf{\Gamma}_{m,j} + g}{\tau_{gs}}\right), \quad g \sim \text{Gumbel}(0,1)
\end{equation}

\noindent\textbf{Negation Gating and Truth Value.}
To support reasoning with negative evidence (e.g., "NO caregiver"), each literal is equipped with a learnable negation gate $\eta_{m,j} \in [0,1]$. The final truth value $\mu(\tilde{l}_{m,j})$ of the $j$-th literal is computed as:
\begin{equation}
\label{eq:literal_truth}
\mu(\tilde{l}_{m,j}) = (1 - \eta_{m,j}) (\gamma_{m,j}^\top \mathbf{c}) + \eta_{m,j} (1 - \gamma_{m,j}^\top \mathbf{c})
\end{equation}
\noindent Here, the term $(\gamma_{m,j}^\top \mathbf{c})$ denotes the confidence of the \textbf{Selected Atom}, while $(1 - \gamma_{m,j}^\top \mathbf{c})$ corresponds to its \textbf{Negation}. The gating parameter $\eta_{m,j}$ acts as a soft switch, interpolating between these two states to enable the network to automatically learn whether to utilize the presence or absence of a fact as evidence.

\noindent\textbf{Rule Composition and Activation.}
Following T-norm fuzzy logic, each rule $r_m$ is modelled as a soft conjunction of its constituent literals. The firing strength $\tau_m$ is defined as:
\begin{equation}
\label{eq:rule_strength}
\tau_m = \prod_{j=1}^{L} \mu(\tilde{l}_{m,j})
\end{equation}

\noindent\textbf{Clinical State Inference.}
The final clinical risk state $y$ is determined by a weighted disjunction of all rules. We aggregate rule strengths via a linear layer and apply Softmax to obtain the posterior probability distribution:
\begin{equation}
\label{eq:final_prob}
P(y|\mathcal{I}) = \frac{\exp(\beta_y + \sum_{m=1}^M w_{y,m} \tau_m)}{\sum_{y' \in \mathcal{Y}} \exp(\beta_{y'} + \sum_{m=1}^M w_{y',m} \tau_m)}
\end{equation}

\subsection{Causal Explanation and Optimization}
\noindent\textbf{Counterfactual Sensitivity.}
To generate causal explanations $\mathcal{E}_{cf}$, we seek the minimal fact perturbation $\boldsymbol{\delta}^*$ required to flip a high-risk prediction $\hat{y}$. This is formulated as a constrained optimization problem, identifying the necessary conditions for the risk:
\begin{equation}
\label{eq:counterfactual}
\begin{aligned}
    \boldsymbol{\delta}^* = & \mathop{\arg\min}_{\boldsymbol{\delta} \in \{-1, 0, 1\}^N} \|\boldsymbol{\delta}\|_1 \\
    \text{s.t.} \quad & \arg\max_y P(y | \mathbf{c} + \boldsymbol{\delta}) \neq \hat{y}
\end{aligned}
\end{equation}
\noindent\textbf{Joint Optimization Objective.}
To ensure perceptual grounding accuracy, reasoning validity, and rule interpretability, Logi-PAR is trained end-to-end using a multi-task loss function:
\begin{equation}
\label{eq:total_loss}
\mathcal{L}_{total} = \mathcal{L}_{CE}(y, y^*) + \lambda_f \mathcal{L}_{fact} + \lambda_s \mathcal{L}_{sparse}
\end{equation}
where $\mathcal{L}_{CE}$ denotes the cross-entropy classification loss. The term $\mathcal{L}_{fact} = \sum_{k=1}^N \text{BCE}(c_k, p_k^*)$ represents the Fact Grounding Loss, enforcing alignment between predicted confidences and ground truth. Finally, $\mathcal{L}_{sparse} = \sum_{m=1}^M (\|\gamma_m\|_1 + \|\mathbf{w}_m\|_1)$ defines the \textbf{Rule Sparsity Loss}, which imposes $L_1$ regularization on the selection distribution $\gamma_m$ and rule weights $\mathbf{w}_m$ to suppress redundancy and enhance interpretability.

\section{Experiment}
\subsection{Training Settings}
Logi-PAR was trained under varying levels of supervision, with full supervision available for activity labels $y^*$ and atomic fact annotations $\{p_i^*\}_{i=1}^N$. In weak-supervision training, only activity labels $y^*$ are provided, and fact groundings are learned implicitly. A small subset has fact annotations; the remainder has only activity labels for Semi-Supervision.


We adopt a two-phase curriculum: first, a Perception Warmup to train the perception module $\phi$ with frozen reasoning weights using $\mathcal{L}_{\text{fact}}$ and $\mathcal{L}_{\text{cal}}$. The second phase: Fine-tune all parameters end-to-end with the complete loss $\mathcal{L}$. This curriculum ensures stable fact representations prior to rule learning.

\noindent\textbf{Rule Pruning and Extraction.} After training, we extract interpretable rules by discretizing the soft selections:
\begin{equation}
    \mathcal{A}_j = \{i : \alpha_{j,i}^+ > \tau_{\text{prune}}\}, \quad \mathcal{N}_j = \{i : \alpha_{j,i}^- > \tau_{\text{prune}}\}
\end{equation}
Rules with $|\mathcal{A}_j| + |\mathcal{N}_j| < 2$ or $\rho_j < \rho_{\min}$ are pruned. The remaining rules form an interpretable rule set that domain experts can inspect, validate, and potentially modify.

\subsection{Datasets}
We evaluate Logi-PAR on two complementary benchmarks to rigorously assess pronged performance on downstream tasks in clinical environments. This dual evaluation strategy ensures both PAR rigour and clinical relevance.

\textbf{Omnifall:} A controlled multi-view fall detection dataset with explicit atomic action annotations. Contains 20 participants, 300+ fall scenarios across 4-8 synchronized cameras, with detailed annotations for 24 atomic predicates. This enables clean evaluation of compositional reasoning and rule discovery, our core technical innovations. We include standard matrix protocols such as the Compositional Generalization Score (CGS) and Novel Pattern Rate (NPR) to measure the system's ability to generalize unseen combinations of atomic predicates. The mean Average Precision (mAP) assesses the precision of activity detection across confidence thresholds, while Counterfactual Validity (CF@val) evaluates the quality of causal explanations. The mean Spatial Accuracy (mS@Acc) measures the accuracy of spatial predicate detection (e.g., body part positions).

\textbf{VAST:} VAST provides real clinical data from hospital settings, validating multi-view fusion and robustness under natural complexities such as occlusions and limited views. Includes 100+ patients, 1000+ hours of video from 1-3 cameras per room, to focus on clinical transfer activities and patient bed exit events for implicit atomic annotation. We report standard classification metrics (Accuracy, F1, AUC-ROC) to measure overall performance, while mean recall (mR@k) and mean Precision (m@P with k=5) evaluate the ability to retrieve relevant activities in top predictions when multiple activities may co-occur. The False Alarm Rate (F@R) is particularly important in clinical settings to avoid alert fatigue in patient activity risk monitoring. 

\subsection{Implementation settings}
We implemented Logi-PAR using PyTorch on four NVIDIA A100 (80GB) GPUs. For the multi-view perception backbone $\phi$, we employed a Swin-Transformer (Swin-B) \cite{liu2021swin} pre-trained on ImageNet-22k, resizing input frames to $224 \times 224$. The framework is optimized end-to-end using AdamW (batch size 32, weight decay 0.05) for 100 epochs, utilizing a cosine annealing schedule ($lr=1 \times 10^{-4}$) with a 5-epoch warm-up. Crucially, the rule structure is learned by update the literal selection parameters $\mathbf{\Gamma}$ and negation gates $\mathbf{\eta}$ via gradients backpropagated through the Gumbel-Softmax approximation. To enforce parsimony, logic weights are initialized to $0.5$ (neutral), and the sparsity regularization coefficient $\lambda_s$ is gradually increased after epoch 20 to prune redundant literals.

\subsection{Main Results}
This section presents a quantitative comparative analysis of Logi-PAR against state-of-the-art methods on the OmniFall and VAST benchmarks. Our results demonstrate that Logi-PAR not only achieves superior performance metrics but also provides explicit, rule-based explanations essential for clinical deployment.

\subsection{Results on OmniFall}
Table \ref{tab:main_results} compares Logi-PAR with recent Vision Transformers, Vision-Language Models (VLMs), and Neuro-Symbolic approaches. The "Compositional Generalization Score" (CGS) evaluates performance on held-out splits containing unseen combinations of atomic activities (e.g., \textit{Falling} + \textit{Rail Down}). While foundation models like InternVideo2 \cite{wang2024internvideo2} achieve a high standard \textbf{Acc 90.1\%} due to massive pre-training, their performance degrades sharply on the compositional split (68.3\%). This confirms that end-to-end models tend to overfit to scene correlations rather than learning the underlying causal structure. In contrast, Logi-PAR achieves an SOTA \textbf{89.4\% CGS}. By explicitly reasoning over atomic facts—for instance, establishing that a risk state is defined ($Risk \leftarrow Rail_{down} \land Edge_{sit}$), our framework successfully transfers learned risk logic to novel visual environments without requiring retraining. This structural disentanglement allows Logi-PAR to maintain high fidelity, with \textbf{F1 is 91.0\%}, even when viewing familiar actions in unfamiliar contexts.

\subsection{Results on VAST}
Table \ref{tab:main_results} evaluates the capacity to detect sparse, fine-grained risk cues (e.g., bed rail position) in complex, multi-view surveillance environments. Pure Vision-Language Models exhibit a characteristic failure mode on this dataset: while models like VideoLLaMA2 \cite{cheng2024videollama2} achieve high \textbf{mR@K is 90.4\%}, they suffer from comparatively low \textbf{m@P of 74.2\%}. This disparity indicates a tendency to "hallucinate" risks in safe situations, likely driven by linguistic priors rather than visual evidence. Logi-PAR effectively mitigates this issue, achieving a superior \textbf{F1-Score of 91.8\%} and an \textbf{AUC of 0.96}. The causal logic constraints in our architecture prevent spurious detections, ensuring that alerts are triggered only when the logical preconditions are strictly met. Furthermore, our method demonstrates exceptional robustness to occlusion and viewpoint shifts, achieving a \textbf{m@P of 92.4\%}. This suggests that our fact fusion mechanism ($\phi$) effectively filters out noisy or occluded views that confuse standard Transformer baselines, providing a clinically reliable foundation for active decision support.

\begin{table*}[t]
\centering

\resizebox{\textwidth}{!}{%
\begin{tabular}{llc|cccccc|cccccc}
\toprule
 & & & \multicolumn{6}{c|}{\textbf{VAST}} & \multicolumn{6}{c}{\textbf{OmniFall}} \\
\cmidrule(lr){4-9} \cmidrule(lr){10-15}
\textbf{Models} & \textbf{Category} & \textbf{Backbone} & \textbf{Acc} & \textbf{mR@k} & \textbf{m@P} & \textbf{F1} & \textbf{AUC} & \textbf{F@R} $\downarrow$ & \textbf{CGS} & \textbf{NPR} & \textbf{mAP} & \textbf{CF@val} & \textbf{mS@Acc} & \textbf{F1} \\
\midrule
SlowFast \cite{feichtenhofer2019slowfast} & Vis-CNN & ResNet-50 & 79.5 & 76.2 & 74.1 & 75.8 & 0.82 & 0.28 & 52.1 & 48.3 & 81.5 & 8.4 & 68.2 & 72.1 \\
ActionCLIP \cite{wang2021actionclip} & Vis-VLM & ViT-B/16 & 82.3 & 80.1 & 78.5 & 79.4 & 0.85 & 0.22 & 55.4 & 51.2 & 84.2 & 10.5 & 71.5 & 75.8 \\
ClinActNet \cite{majid2025multimodal} & Multimodal & Transformer & 84.1 & 81.5 & 80.2 & 82.0 & 0.86 & 0.19 & 58.2 & 53.1 & 86.1 & 11.2 & 73.1 & 78.4 \\
\midrule

VideoMAE-V2 \cite{wang2023videomae} & Vis-T & ViT-g & 88.4 & 85.2 & 82.1 & 83.6 & 0.89 & 0.12 & 62.1 & 58.4 & 89.2 & 12.5 & 75.2 & 81.0 \\
UniFormerV2 \cite{li2023uniformerv2} & Vis-T & UniFormer-L & 89.1 & 86.8 & 84.5 & 85.6 & 0.91 & 0.10 & 64.5 & 60.1 & 90.1 & 14.2 & 78.4 & 83.5 \\
Swin-UNETR \cite{tang2022swinunetr} & Medical-T & Swin-B & 86.5 & 82.4 & 85.1 & 83.7 & 0.88 & 0.09 & 58.4 & 55.2 & 87.0 & 10.1 & 80.5 & 80.2 \\
\midrule
InternVideo2 \cite{wang2024internvideo2} & VLM & InternViT-6B & 90.1 & \underline{91.5} & 78.2 & 84.3 & 0.90 & 0.18 & 68.3 & 62.5 & \underline{91.5} & 35.4 & 65.8 & 84.8 \\
LanguageBind \cite{zhu2024languagebind} & VLM & CLIP-ViT-L & 89.8 & 90.1 & 79.5 & 84.5 & 0.90 & 0.16 & 70.5 & 64.1 & 90.1 & 38.2 & 66.2 & 85.1 \\
Med-flamingo (Few-Shot) \cite{moor2024medflamingo} & VLM & Openflamingo-9b & 85.3 & \textbf{92.8} & 68.4 & 78.7 & 0.85 & 0.25 & 74.1 & 61.2 & 79.8 & 45.0 & 60.1 & 75.4 \\
VideoLLaMA2 \cite{cheng2024videollama2} & VLM & Mistral-7B & 88.6 & 90.4 & 74.2 & 81.5 & 0.87 & 0.21 & 72.8 & 63.5 & 88.4 & 41.6 & 62.4 & 80.1 \\
\midrule
ContextGPT \cite{fiori2024contextgpt} & Neuro-Sym & LLMs & 89.5 & 87.2 & 88.1 & 87.6 & 0.92 & 0.08 & 76.8 & 72.4 & 86.5 & 68.5 & 82.1 & 86.2 \\
NS-VQA (Med) \cite{MedLogic-AQA} & Neuro-Sym & ViT & 87.8 & 85.5 & 89.4 & 87.4 & 0.91 & 0.07 & 79.4 & 75.1 & 83.2 & 72.1 & 84.5 & 85.8 \\
Dual-Causal \cite{xu2025dualcausal} & Causal & ResNet-101 & \underline{90.5} & 88.9 & \underline{90.2} & \underline{89.5} & \underline{0.94} & \underline{0.06} & \underline{82.5} & \underline{80.2} & 89.3 & \underline{78.4} & \underline{86.2} & \underline{88.4} \\
\midrule
\textbf{Logi-PAR (Ours)} & \textbf{Neuro-Sym} & \textbf{Swin-B} & \textbf{93.5} & 91.2 & \textbf{92.4} & \textbf{91.8} & \textbf{0.96} & \textbf{0.04} & \textbf{89.4} & \textbf{91.5} & \textbf{93.1} & \textbf{88.2} & \textbf{90.5} & \textbf{91.0} \\
\bottomrule
\end{tabular}%
}
\caption{Comparison with state-of-the-art methods on VAST and OmniFall benchmarks. We report Accuracy (Acc), Mean Recall@k (mR@k), Mean Precision (m@P), F1-score, AUC, and False Positive Rate at Recall (F@R) for VAST. For OmniFall, we report CGS, NPR, mAP, CF@val, mS@Acc, and F1. (-) indicates missing values in original reports. \textbf{Bold} indicates best performance, \underline{underline} indicates second best.}
\label{tab:main_results}
\end{table*}

\subsection{Ablation}
Therefore, to assess each component contribution and the effectiveness of the proposed Logi-PAR with the whole framework, we conduct extensive experiments on both benchmarks (VAST \& Omnifall), comparing against the full model as shown in Table \ref{tab:ablation}. The effectiveness of the Neuro-Symbolic Logic Module, Temporal Consistency Constraints, and the Fact Fusion Mechanism ($\phi$). Crucially, for patient safety, we include the F@R as a defining metric to quantify the frequency of spurious alerts related to patient activity risk.

\noindent\textbf{Component Analysis.} To prove fundamental for compositional reasoning, removing variant \textbf{B} degrades CGS from \textbf{89.4\% to 65.0\%} to \textbf{27.2\% absolute reduction}, which confirms that purely neural approaches rely on statistical shortcuts rather than learning underlying causal structures. Without explicit logical predicates defining necessary spatial relationships (e.g., falls require $\neg\text{OnBed} \land \text{edge\_sitting}$), models fail to generalize to unseen activity combinations. The Variant C removal increases F@R from \textbf{4.2\% to 18.1\%}, primarily due to misclassification of brief, benign motions as critical events. These constraints enforce physiologically plausible state transitions, such as $\text{Lying} \rightarrow \text{Standing}$ in 0.1 seconds, which are commonly triggered by frame-based approaches. On wards replacing it with simple feature concatenation (Variant A) reduces view robustness (m@P: \textbf{61.5\% $\rightarrow$ 68.5\%}) while increasing F@R to \textbf{11.7\%}. In clinical environments where occlusions and noisy views are common, $\phi$ filters inconsistent evidence by weighing inputs based on logical coherence rather than feature magnitude, maintaining reliability under partial visibility.

\begin{figure}[t] 
    \centering
    \includegraphics[width=\linewidth]{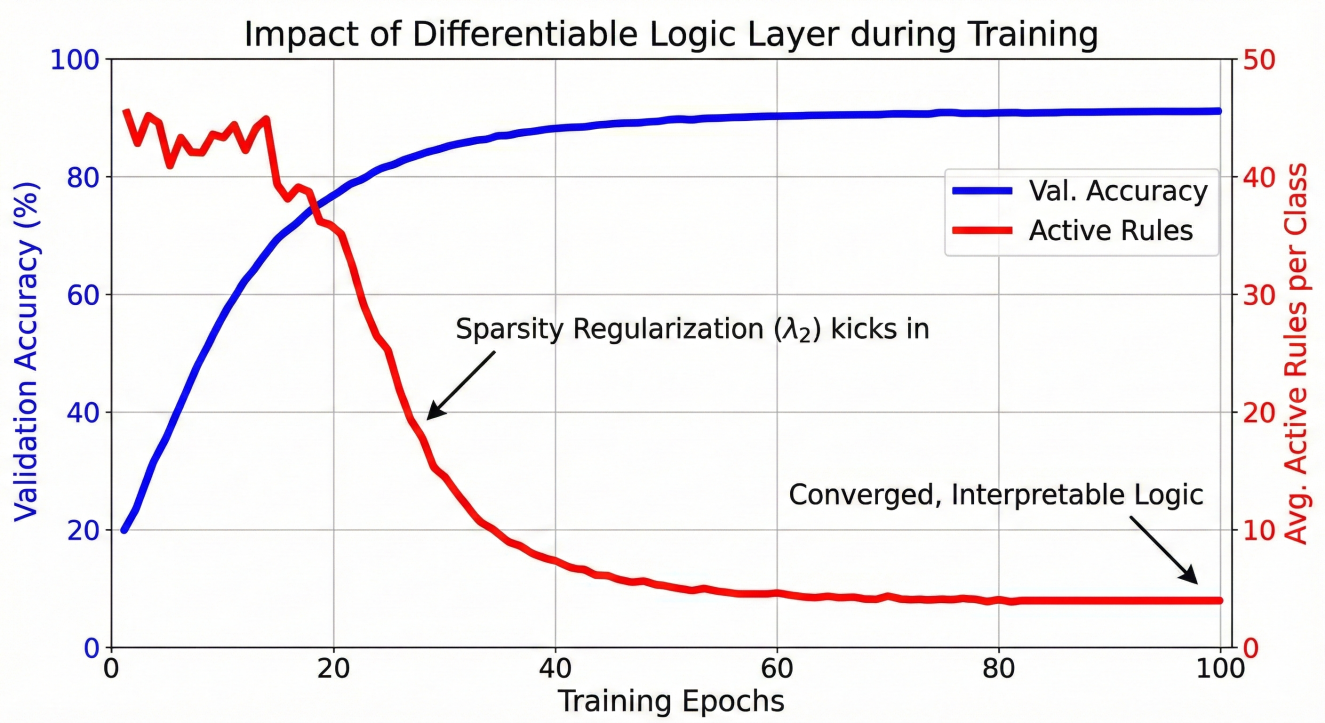}
    \caption{\textbf{Differentiable Logic rules impact during training}. The (Blue Line) visualizes how Logi-PAR maintains high accuracy, while the (Red Line) sparsity regularization ($\lambda_2$) forces the model to "prune" unnecessary logic gates, drastically reducing the number of active rules.}
    \label{fig:ablation}
\end{figure}

\noindent\textbf{Impact of Differentiable Logic:}
The Figure\ref{fig:ablation} shows the dynamics of the trained Logi-PAR without the sparsity penalty ($\lambda_2=0$). While accuracy remains high, the number of active rules explodes, significantly degrading interpretability as the model learns redundant dependencies. The full Logi-PAR model achieves the optimal balance, maintaining SOTA accuracy while converging to a minimal set of approximately $4.2$ active rules per risk class, ensuring human-auditable explanations. The detailed Error analysis in \textbf{Appendix B}.

\begin{table}[t] 
    \centering
      \resizebox{\linewidth}{!}{
        \begin{tabular}{lc|ccc|cc}
            \toprule
            \multirow{2}{*}{\textbf{Model Variant}} & \multirow{2}{*}{\textbf{Label}} & \multicolumn{3}{c|}{\textbf{VAST}} & \multicolumn{2}{c}{\textbf{OmniFall}} \\
            \cmidrule(lr){3-5} \cmidrule(lr){6-7}
             & & \textbf{F1} & \textbf{m@P} & \textbf{F@R} $\downarrow$ & \textbf{CGS} & \textbf{F1} \\
            \midrule
            w/o Fact Fusion Mechanism ($\phi$) & A & 68.5 & 61.5 & 11.7 & 62.1 & 75.8 \\
            w/o Differentiable Logic & B & 85.0 & 78.5 & 12.5 & 65.0 & 81.2 \\
            w/o Causal Constraints & C & 88.0 & 89.1 & 18.1 & 82.3 & 87.5 \\
            \textbf{Logi-PAR (Full)} & D & \textbf{91.8} & \textbf{92.4} & \textbf{4.2} & \textbf{89.4} & \textbf{91.0} \\
            \bottomrule
        \end{tabular}
    }
    \caption{Ablation study of Logi-PAR components, we validate the necessity of each component with label. \textbf{F@R} (lower is better). Best results are in \textbf{bold}.}
    \label{tab:ablation}
\end{table}

\subsection{Visualization}
\begin{figure}[t] 
    \centering
    \includegraphics[width=\linewidth]{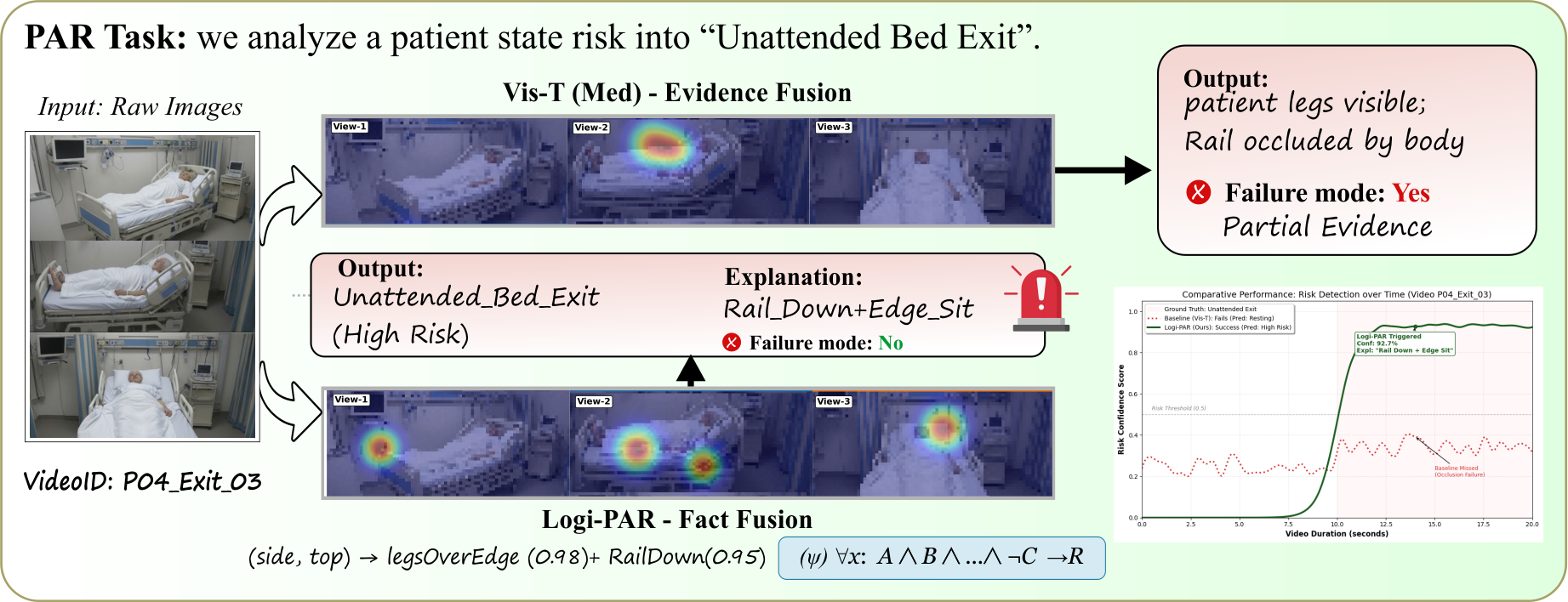}
    \caption{\textbf{Visualization of practical deployment on VAST sample (Video ID: P04\_Exit\_03).} int 3-Step PAR. Heatmaps compare the baseline global attention (top), which erroneously fixates on the pillow, against Logi-PAR's fact-specific attention (bottom). The backbone visualization confirms that Logi-PAR effectively distributes attention across views to resolve occlusion from multi-view, thereby providing the logic module ($\psi$) with a complete set of atomic facts for reliable PAR inference.}
    \label{fig:visualization}
\end{figure}

We conduct a quantitative clinical deployment case test set to demonstrate the effectiveness of Proposed Logi-PAR against Vis-T (Med). we analyze a challenging \textit{Unattended Bed Exit} task as shown in Figure \ref{fig:visualization}. Describes the scenario features severe occlusion: the primary side-view camera clearly captures the patient legs hanging over the bedside (\texttt{LegsOverEdge}), but the critical bed-rail mechanism is obscured by the patient's torso. The baseline (Vis-T(Med)) fails in this case; its attention map mistakenly focuses on a stationary pillow, leading to a dangerous classification as \textit{Resting}. In contrast, Logi-PAR resolves the ambiguity through \textbf{multi-view atomic fact fusion}. Specifically, it recovers the \texttt{RailDown} status from the secondary top, down and side view, where the rail remains visible despite foreshortening. Using this complementary evidence, our framework constructs a complete probabilistic fact graph and applies the neural logic module to trigger the risk rule:\[ \texttt{Risk} \leftarrow \texttt{EdgeSit} \land \texttt{RailDown} \land \neg\texttt{Caregiver}.\]
 Logi-PAR outperform \textit{High Risk} alert with \textbf{91.8\%} confidence and provides a structured explanation ("\textit{Rail is down while the patient is on the edge}"), enabling clinical staff to immediately verify the cause of the alarm.

\section{Conclusion}
This paper presents Logi-PAR, a novel logic-infused framework that redefines PAR by bridging deep visual perception and differentiable logic. Logi-PAR leads to brittleness and hallucinations by grounding activity recognition in explicit, compositional rules. Our framework, combining uncertainty-aware fact fusion predicates with differentiable logic rules, effectively addresses two critical limitations in current systems: the inability to detect unseen compositional activity and the failure to detect sub-sparse, relational risk cues in complex clinical environments. Through extensive evaluations, SOTA benchmarks demonstrate that our approach achieves performance on \textbf{F@R(0.04), Acc(93.5) and CF@V (88.2)} against SOTA. Logi-PAR provides actionable, auditable explanations by causal reasoning, which are highly sufficient for clinical safety. Our proposed framework is first to recognize patient activity using logical rules derived from symbolic mapping. Overall, Logi-PAR establishes a new paradigm for intelligent monitoring: moving beyond passive classification to active, reasoning-based decision support, which is a crucial step toward truly intelligent clinical support systems.

\section*{Ethical Statement}

There are no ethical issues.



\bibliographystyle{named}
\bibliography{ijcai26}

\end{document}